\documentclass[lettersize,journal]{IEEEtran}
\usepackage{amsmath,amsfonts}
\usepackage{algorithmic}
\usepackage{algorithm}
\usepackage{array}
\usepackage[caption=false,font=normalsize,labelfont=sf,textfont=sf]{subfig}
\usepackage{textcomp}
\usepackage{stfloats}
\usepackage{url}
\usepackage{verbatim}
\usepackage{graphicx}
\usepackage{hyperref}
\usepackage{cite}
\usepackage{multirow}
\usepackage{threeparttable}
\usepackage{booktabs}
\usepackage{color} 
\usepackage{amssymb}
\usepackage{amsmath}
\usepackage{diagbox}
\usepackage{utfsym}
\begin{document}

\title{Bidirectional skip-frame prediction for video anomaly detection with intra-domain disparity-driven attention}

\author{\IEEEauthorblockN{Jiahao Lyu\textsuperscript{1},Minghua Zhao\textsuperscript{1,\dag},Jing Hu\textsuperscript{1}, Runtao Xi\textsuperscript{2}, Xuewen Huang\textsuperscript{1}, Shuangli Du\textsuperscript{1}, Cheng Shi\textsuperscript{1}, Tian Ma\textsuperscript{3}}
	\\
	\IEEEauthorblockA{\textit{1. School of Computer Science and Engineering, Xi'an University
			of Technology, Xi'an, China}
		\\
		\IEEEauthorblockA{\textit{2. CCTEG Changzhou Research Institute;Tiandi(Changzhou) Automation Co., Ltd.}}	
		\\
		\IEEEauthorblockA{\textit{3. College of Computer Science and Technology, Xi'an University of Science and Technology}}
		\\	
		\IEEEauthorblockA{\dag	Correspondence:\href{mailto:zhaominghua@xaut.edu.cn}{zhaominghua@xaut.edu.cn}}
	}
}

\maketitle

\begin{abstract}
With the widespread deployment of video surveillance devices and the demand for intelligent system development, video anomaly detection (VAD) has become an important part of constructing intelligent surveillance systems. Expanding the discriminative boundary between normal and abnormal events to enhance performance is the common goal and challenge of VAD. To address this problem, we propose a Bidirectional Skip-frame Prediction (BiSP) network based on a dual-stream autoencoder, from the perspective of learning the intra-domain disparity between different features. The BiSP skips frames in the training phase to achieve the forward and backward frame prediction respectively, and in the testing phase, it utilizes bidirectional consecutive frames to co-predict the same intermediate frames, thus expanding the degree of disparity between normal and abnormal events. The BiSP designs the variance channel attention and context spatial attention from the perspectives of movement patterns and object scales, respectively, thus ensuring the maximization of the disparity between normal and abnormal in the feature extraction and delivery with different dimensions. Extensive experiments from four benchmark datasets demonstrate the effectiveness of the proposed BiSP, which substantially outperforms state-of-the-art competing methods.
\end{abstract}


\section{Introduction}

With the development of computer vision, anomaly detection has been widely researched in many fields and applications, such as medical image anomaly detection \cite{tian2021constrained}, industrial quality inspection \cite{fang2023fastrecon,roth2022towards}, and traffic surveillance \cite{cheng2023spatial,gong2019memorizing}. In particular, with the widespread popularization and application of video surveillance equipment in different fields of society, it plays an important role in guaranteeing social security and stability. Simultaneously, the quantity of surveillance videos has shown explosive growth, and manual detection methods cannot meet the demand for timely and accurate detection of massive video. Therefore, intelligent technology is gradually replacing manual detection, among that, video anomaly detection (VAD) has been investigated and studied by researchers with the development of computer vision.

\begin{figure}[ht] 
	\centering
	\includegraphics[width=\linewidth]{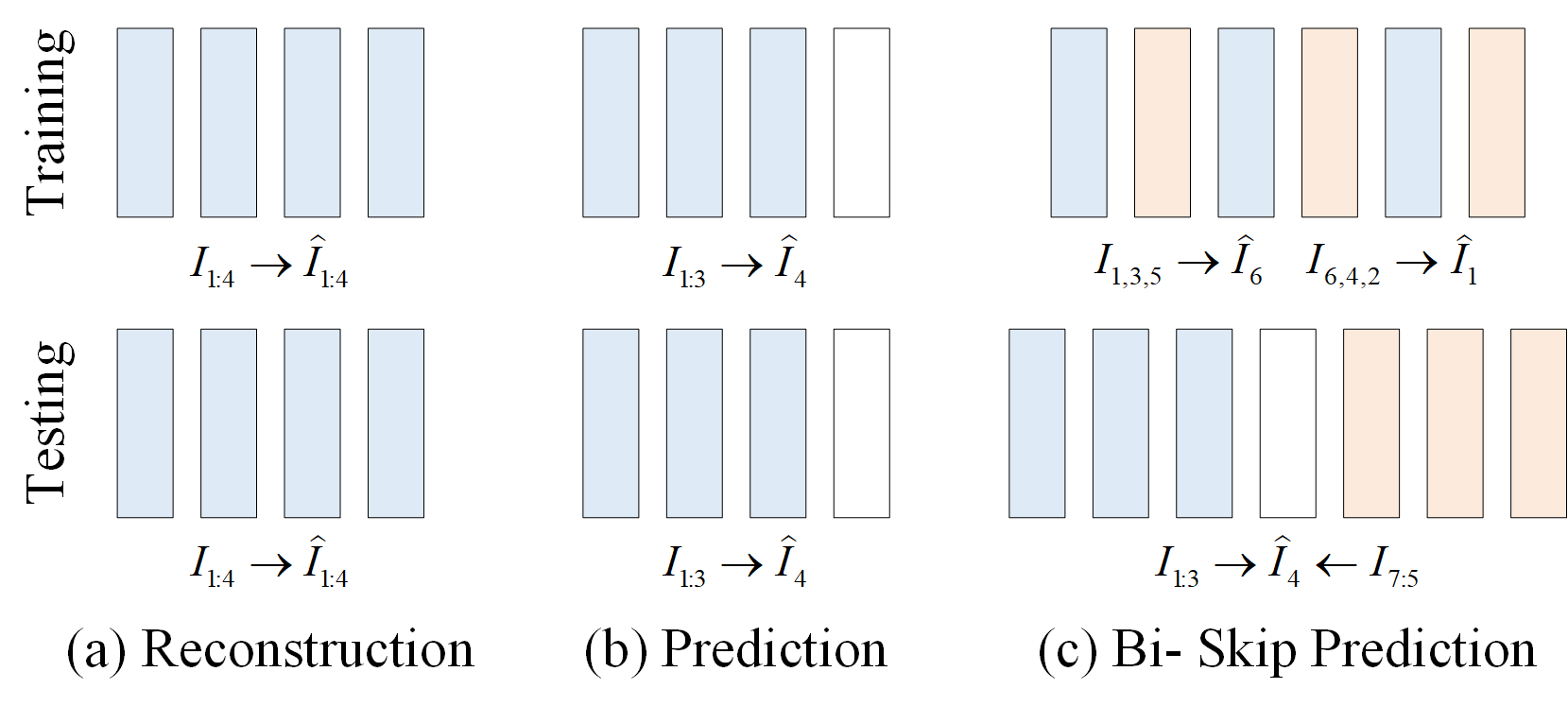}
	\caption{Three types of video anomaly detection methods. (a) Reconstruction-based method. (b) Prediction-based method. (c) The proposed Bidirectional Skip-frame Prediction method.} \label{fig1}
\end{figure}

VAD can be categorized into supervised learning \cite{ionescu2019object}, weakly-supervised learning \cite{zeng2023temporal}, and unsupervised learning \cite{gong2019memorizing}. Supervised VAD methods are not considered the major method because they need to utilize the fine-grained label or image-level label data to complete the discrimination between normal and anomaly. The anomaly events are usually discrete and small probability events in the whole video sequence, which is a very time-consuming and inefficient method. Similarly, weakly-supervised VAD methods utilize image-level labels to make anomaly detection, and the training dataset consists of labeled normal and anomaly events, to identify between normal and anomaly through multi-instance learning ranking \cite{feng2021mist,liu2022abnormal} or classification models \cite{majhi2024oe}. Compared to the above two types of methods, unsupervised VAD methods do not need to provide any labels and only need to complete the classification preprocessing of normal and abnormal events when constructing the dataset, i.e., only modeling and feature learning of normal events in the training phase, and completing the anomaly detection by measuring the reconstruction deviation of the abnormal events in the testing phase, which can simply locate the specific temporal frames and spatial locations of the abnormal events. Unsupervised VAD methods all follow the generalized assumption \cite{pang2021deep} for anomaly detection tasks, i.e., all events that did not occur in the training set are considered anomalies. Therefore, the boundary of discrimination between normal and abnormal events is particularly important, but there is a serious challenge with this assumption, i.e., when the boundary is ambiguous \cite{huang2022pixel}, the anomaly detection model will suffer tremendously, leading to performance degradation. To this end, previous works have investigated from the perspectives of spatio-temporal feature extraction \cite{hao2022spatiotemporal,park2022fastano} and attention mechanisms \cite{zhang2022hybrid}, among which, most of the methods implement reconstruction-based or prediction-based methods with the Autoencoder (AE) framework, where the encoder compresses the input features into low-dimensional features, and the decoder reduces the compressed features to the input features as much as possible, through the reconstruction or prediction errors to complete VAD. Compared to reconstruction-based methods, prediction-based methods have been widely researched as they can model both temporal and spatial features. As shown in Fig.~\ref{fig1}(a)(b), reconstruction-based methods input and output are the same frames, i.e., input continuous frames $I_{1:4}$, output reconstructed frames $\hat{I}_{1:4}$. Whereas the prediction-based methods input $I_{1:3}$, output the predicted frame $\hat{I}_{4}$.

In the current reconstruction \cite{gong2019memorizing,chang2020clustering} or prediction methods \cite{cheng2023spatial,liu2018future}, the anomaly is determined by calculating the reconstruction/prediction error between the original frame and the reconstructed/predicted frame. For normal events, the reconstruction/prediction error needs to be small, while for abnormal events, the reconstruction/prediction error needs to be large. Therefore, how to minimize the normal errors and maximize the anomaly errors is the essential problem of VAD. However, most of the methods only determine the detection accuracy by a single type of error. To enhance the comprehensiveness of the detection and reduce the false alarm rate, some works have proposed hybrid anomaly detection \cite{liu2021hybrid,wang2023memory,liang2024c,singh2024attention} or bidirectional prediction  \cite{fang2020anomaly,zhong2022bidirectional,li2023multi} methods, such as fusing the frame error with the optical flow error, fusing the reconstruction error with the prediction error, and fusing the forward prediction error with the backward prediction error, etc., which can help to improve the detection performance from different perspectives. Therefore, some methods utilize preprocessing operations such as the optical flow extraction \cite{liu2021hybrid} or the foreground mask \cite{luo2021future} as auxiliary information, but the additional time consumption makes them unable for real-time tasks.

Due to the weak intra-domain gap between normal events and the large intra-domain disparity between normal and anomaly events, the VAD method allows normal events to form clusters in the feature space, while most abnormal features cannot belong to the same cluster with them, or even form separate abnormal clusters. Inspired by this, and follow the bidirectional prediction method \cite{fang2020anomaly,zhong2022bidirectional}, this paper design a bidirectional skip-frame prediction (BiSP) method, as shown in Fig.~\ref{fig1}(c). Unlike traditional single-frame \cite{liu2018future} and dual-frame \cite{fang2020anomaly} prediction methods, the proposed training strategy uses skip frames to predict future frames at the end of the bidirectional direction. For instance, using forward skip frames $I_{1,3,5}$ to predict $\hat{I}_{6}$ and using backward skip frames $I_{6,4,2}$ to predict $\hat{I}_{1}$. Compared with the traditional prediction method, the proposed method can extract the motion features more easily and focuses on the normal events with the weak intra-domain gap by the proposed attention, and then change the prediction mode in the testing phase, to increase the intra-domain disparity between the normal and abnormal events further. Ultimately, we realize intra-domain disparity-driven VAD for surveillance videos. Specifically, based on the dual-stream AE framework, we construct a variance channel attention with parallel structure and a context spatial attention with serial structure, which enhance the model's ability to discriminate the normality of features of different dimensions in the skip connection and decoding processes, respectively. Specifically, on the one hand, variance channel attention enhances the ability of the model to discriminate movement patterns between different events.  On the other hand, context spatial attention enriches the feature extraction and representation ability of the model for objects with different scales. In the testing phase, consecutive frames at both ends of the input are used to co-predict the same intermediate frames, and the disparity between normal and abnormal events is enlarged compared to the weak gap between normal and normal events. 

In brief, the proposed method considers how to use the attention to expand the discrimination boundary between normal and abnormal events in both forward and backward prediction, and finally achieve efficient detection by maximum anomaly error. In summary, the main contributions of this paper are as follows.

\begin{itemize}
	\item We propose a novel bidirectional skip-frame prediction method that utilizes the dual-AE structure for both forward and backward prediction. Compared to state-of-the-art methods, it achieves competitive performance on benchmark datasets.
	
	\item Due to the intra-domain disparity between normal and anomaly events, the proposed method can further expand the disparity by adopting different video frame input strategies in the training and testing stages.
	
	\item To enhance feature representation, we propose variance channel attention and context spatial attention respectively. The purpose of the two attention is to better discriminate different categories of anomaly from the perspective of the anomaly's movement pattern and appearance, respectively.
	
\end{itemize}

The rest of this paper is organized as follows: Section~\ref{sec2} discusses the related work for VAD, Section~\ref{sec3} introduces the framework of the proposed BiSP, Section~\ref{sec4} presents the experimental and visualization results on the benchmark dataset, and Section~\ref{sec5} concludes the paper.

\section{Related Work} \label{sec2}

The majority of video clips captured by surveillance equipment depict normal events, rendering abnormal events even rarer and harder to define. Therefore, it is more reliable to use unsupervised learning for VAD. Most unsupervised VADs are mainly categorized into reconstruction-based \cite{gong2019memorizing} and prediction-based \cite{liu2018future}, both of which discriminate abnormal events by measuring the error in the reconstructed or predicted frames. The difference is that reconstruction-based methods identify errors from multiple consecutive frames corresponding to the input, while prediction-based methods focus more on the errors of future frames. Most of these prediction methods predict a single frame for the next moment, but a few methods  \cite{fang2020anomaly,li2023multi,zhang2022mutual} still choose to accomplish a dual-frame prediction.

\noindent\textbf{Reconstruction-based Methods.} Reconstruction methods focus on reconstructing the spatial features of the normal samples due to the input and output of the frames at the same time. For instance, Gong \emph{et al.} \cite{gong2019memorizing} proposed a memory-augmented AutoEncoder (memAE), which embeds an external memory network between the encoder and decoder to record the feature distributions of normal events for efficient reconstruction. Lappas \emph{et al.} \cite{lappas2024dynamic} introduced Dynamic Distinction Learning (DDL) to enhance the accuracy of VAD through the fusion of pseudo anomaly and dynamic anomaly weighting. Kommanduri \emph{et al.} \cite{kommanduri2024dast} introduce the DenserResNet AutoEncoder, which utilizes multiple dense and residual networks to enhance contextual understanding of features across different scales. Park \emph{et al.} \cite{park2020learning} proposed an attention-based memory addressing mechanism, called MNAD, and the testing phase to update the memory pool to ensure that the model can identify normal events. In the same framework, MNAD based on prediction performs much better than reconstruction, which has been similarly found in the paper \cite{qiu2024video}. This is because reconstruction cannot reason in the temporal feature dimension and ignores the temporal relationship between frames.

\noindent\textbf{Prediction-based Methods.} In contrast, the input frames of prediction methods are different from the output frames, which allows better modeling and discovery of temporal relationships in consecutive frames. Frame-Pred proposed by Liu \emph{et al.} \cite{liu2018future} is the first prediction framework to accomplish the VAD task. Huang \emph{et al.} \cite{huang2022self},  in order to reduce the generalization ability of the model to anomaly frames, propose a novel approach to producing rotation-detectable for normal frames, which improves the discriminative ability of the self-supervised discriminator. Ning \emph{et al.} \cite{ning2024memory} proposed the Memory-Augmented Appearance-Motion Consistency framework and applied channel attention to multiple scale features to capture the interaction between appearance and motion (optical flow) information, which finally enhanced the robustness and effectiveness of the model. Huang \emph{et al.} \cite{huang2022boosting} proposes a novel variational normal inference (VNI) model to evaluate the distribution of potential features of normal events, as well as a marginal learning embedding module to optimize VNI model for training. It is shown to remain able to achieve good performance in cross-validation datasets. Compared to 2D convolution, 3D convolution enables joint extraction of temporal and spatial features. Park \emph{et al.} \cite{park2022fastano} designed a spatio-temporal patch transformation model, where consecutive adjacent frames are chopped up in the training phase and made into a frame rectangle as input and randomly generates patch anomalies within the rectangle, which facilitates feature learning. Hao \emph{et al.} \cite{hao2022spatiotemporal} proposed a spatio-temporal coherence enhancement network that extracts and fuses motion and image appearance through 3D-2D convolutional network architecture, to extract spatio-temporal high-level features through 3D convolution during the encoding process, and then realizing feature downscaling by utilizing resampling, and ultimately completing prediction in the decoding process through 2D convolution to accomplish prediction. Qiu \emph{et al.} \cite{qiu2024video} enhanced the ability of the model to learn spatio-temporal features by extracting I3D features, and designed a dual-scale feature cluster module at latent space to expand the discriminative boundary between normal and abnormal events. However, the 3D convolution in the above methods is computationally expensive and cannot guarantee detection efficiency.

Furthermore, hybrid detection and dual-frame prediction methods have been shown to provide better anomaly detection performance. MAAM-Net \cite{wang2023memory} is based on memory augmented module for single frame reconstruction and optical flow prediction, which realizes the fusion error detection in two types of tasks. Fang \emph{et al.} \cite{fang2020anomaly} proposed SIamese generative network (SIGnet), which transforms anomaly detection into a mutually supervised problem, using the two same generators to complete the prediction of the same video frame between two video frames, and based on the prediction consistency loss function, thus improving the model generalization ability. Li \emph{et al.} \cite{li2023multi} proposed a multi-branch generative adversarial network with context learning (MGAN-CL) method and then learned the video frame context information to determine whether abnormal events occur. Zhang \emph{et al.} \cite{zhang2022mutual} utilizes mutual learning and inspiration between different results by using two generators with the same structure and different initialization to predict the same future frames so that they learn from each other, indirectly increasing the diversity of the training samples. This avoids the model overfitting to normal events and enhances the accuracy of anomaly detection. Zhong \emph{et al.} \cite{zhong2022bidirectional} based on bidirectional prediction with the CBAM \cite{woo2018cbam} model, which enables the forward and backward prediction models to achieve high-quality bidirectional detection by adaptively fusing bidirectional spatio-temporal features, respectively.

Compared to reconstruction-based methods, prediction-based methods achieve better detection performance by well-modeling temporal features in the samples. Such methods can better capture the dynamic changes of the samples by considering the temporal and spatial relationships of different events. In contrast, reconstruction-based methods, which cannot consider the temporal relationships among the samples, may fail to recognize normal and abnormal events in small probability cases, thus reconstructing both events at the same time. Therefore, inspired by the intra-domain disparity between normal and abnormal events, this paper proposes a new prediction method with two attention mechanisms.

\begin{figure*}[ht]
	\centering
	\includegraphics[width=\textwidth]{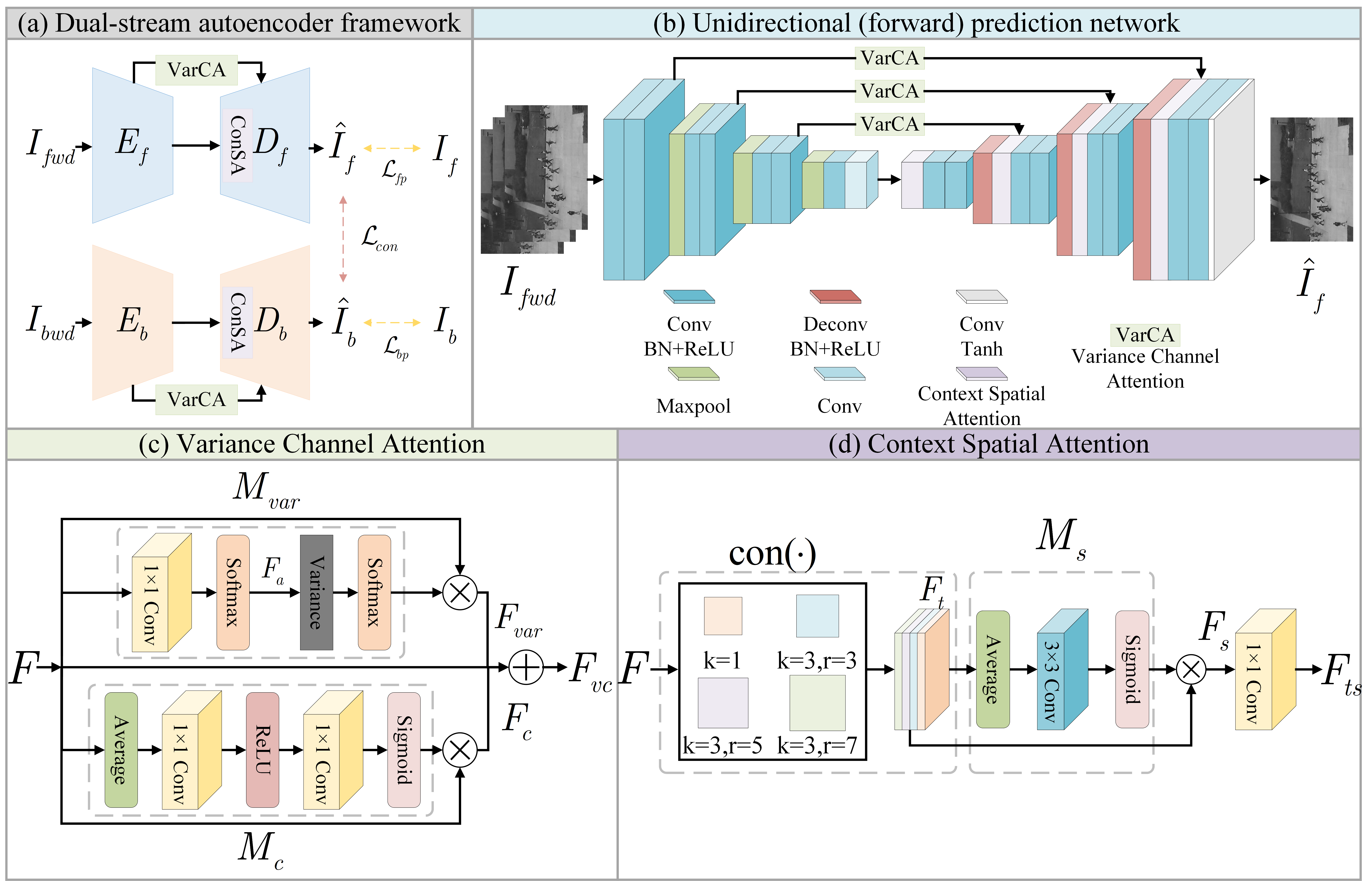}
	\caption{Overview of the proposed Bidirectional Skip-frame Prediction Framework. (a) We design two independent AEs to extract forward and backward skip frame features and predict the next frame respectively. (b) The dual AE have the same structure, so the forward prediction network is used as an example. (c) The parallel structure for variance channel attention. (d) The serial structure for context spatial attention.} \label{fig2}
\end{figure*}

\section{Method} \label{sec3}

In this section, we propose a new unsupervised anomaly detection prediction method and analyze the objective of the proposed BiSP. Meanwhile, the variance channel attention and context spatial attention mechanisms are designed to enhance the model's ability to determine the disparity between normal and anomaly events during different feature delivery processes, respectively. The framework of our proposed BiSP is shown in Fig.~\ref{fig2}(a).

\subsection{Bidirectional Skip-frame Prediction Framework}
From the perspective of expanding the intra-domain disparity between normal and abnormal events, we propose the BiSP, which consists of three modules: dual-stream AE, variance channel attention (VarCA), and context spatial attention (ConSA). Fig.~\ref{fig2}(b) shows the forward frame prediction network in BiSP, which first selects a continuous video frame and then divides it into two separate skip frames, i.e., forward skip frames $I_{fwd}$ and backward skip frames $I_{bwd}$, and then inputs the two skip frames into the corresponding forward $P_{f}=\left \{ E_{f},D_{f} \right \} $ and backward $P_{b}=\left \{ E_{b},D_{b} \right \} $ AE to construct the bidirectional skip-frame prediction network, where $E$ and $D$ denote the encoder and decoder, respectively. The dual AE are identical in structure and differ only in extracting forward or backward temporal features. Forward prediction network $P_{f}(I_{fwd})\rightarrow \hat{I}_{f} $: input forward skip frames $I_{fwd}$, output forward predicted frame $\hat{I}_{f}$. Correspondingly, the backward prediction network $P_{b}(I_{bwd})\rightarrow \hat{I}_{b} $: input backward skip frames $I_{bwd}$, output backward predicted frame $\hat{I}_{b}$. Table~\ref{tab0} shows the detailed composition and specific parameters of the encoders and decoders in the single AE network structure, VarCA and ConSA will be presented in the subsequent section.

\begin{table}[ht]
	\centering
	\caption{The detail of single AE network structure.}\label{tab0}
	\begin{tabular}{l|l|l|l}
		\hline
		Stage & Layers & Kernel size & Stride  \\
		\hline
		\multirow{8}{*}{Encoder} & [Conv+BN+ReLU] $\times 2$ & $3\times  3 \times 32$ & 1 \\
		& [Maxpool] & $ 2\times 2 \times 32$ & 2 \\
		& [Conv+BN+ReLU] $\times 2$ & $3\times  3 \times 64$ & 1  \\
		& [Maxpool] & $ 2\times  2 \times 64$ & 2 \\
		& [Conv+BN+ReLU] $\times 2$ & $3\times  3 \times 128$ & 1  \\
		& [Maxpool] & $ 2\times  2 \times 128$ & 2 \\
		& [Conv+BN+ReLU] & $3\times  3 \times 256$ & 1  \\
		& [Conv] & $3\times  3 \times 256$ & 1   \\
		\hline
		\multirow{8}{*}{Decoder} & [Conv+BN+ReLU] $\times 2$ & $3\times  3 \times 256$ & 1  \\
		& [DeConv+BN+ReLU] & $ 3\times 3 \times 128$ & 2  \\
		& [Conv+BN+ReLU] $\times 2$ & $3\times  3 \times 128$ & 1  \\
		& [DeConv+BN+ReLU] & $ 3\times 3 \times 64$ & 2  \\
		& [Conv+BN+ReLU] $\times 2$ & $3\times  3 \times 64$ & 1  \\
		& [DeConv+BN+ReLU] & $ 3\times 3 \times 32$ & 2  \\
		& [Conv+BN+ReLU] $\times 2$ & $3\times  3 \times 32$ & 1  \\
		& [Conv+Tanh] &  $3\times  3 \times 3$ & 1   \\
		\hline
	\end{tabular}
\end{table}

BiSP aims to complete the bidirectional prediction of frame segments based on bidirectional skip features. To this end, the BiSP uses the L2 distance to compute the corresponding ground truth errors as forward prediction loss $\mathcal{L}_{fp}$ and backward prediction loss $\mathcal{L}_{bp}$:

\begin{equation}\label{eq1}
\mathcal{L}_{fp}=\left\|\hat{I}_{f}-I_{f}\right\|_{2}^{2},
\end{equation} 

\begin{equation}\label{eq2}
\mathcal{L}_{bp}=\left\|\hat{I}_{b}-I_{b}\right\|_{2}^{2}.
\end{equation}

Due to the different prediction goals in the training and testing phases, to ensure the structural consistency of the two final prediction frames in the testing phase, we introduce the Structural Similarity Index Measure (SSIM) to construct the consistency loss $\mathcal{L}_{con}$ of the two prediction frames:

\begin{equation}\label{eq3}
\mathcal{L}_{con}=1-\operatorname{SSIM}(\hat{I}_{f}, \hat{I}_{b}).
\end{equation}

Ultimately, the combination of loss function including two prediction losses and consistency loss is shown in Eq.~(\ref{eq4}):

\begin{equation}\label{eq4}
\mathcal{L}=\mathcal{L}_{fp}+\mathcal{L}_{bp}+\mathcal{L}_{con}.
\end{equation}

VADs usually use anomaly scores to determine whether each video frame is abnormal or not, and most of the methods use peak signal-to-noise ratio (PSNR) to measure the anomaly scores $S(I_{t})$, and PSNR is calculated by the mean-square error $e(I_{t}, \hat{I}_{t})$ between the ground truth $I_{t}$ and the predicted frame $\hat{I}_{t}$. Meanwhile, considering that in this work, two identical frames are predicted in the testing phase, we thus fused the forward $e_{f}(I_{f}, \hat{I}_{f})$ and backward $e_{f}(I_{b}, \hat{I}_{b})$ errors by the weighted sum strategy in Eq.~(\ref{eq6.1}):

\begin{equation}\label{eq6.1}
e(I_{t}, \hat{I}_{t})=w_{f}\cdot e_{f} + w_{b}\cdot e_{b}
\end{equation}
where $w_{f}$ and $w_{b}$ are the weights of the $e_{f}$ and $e_{b}$ scores respectively. Because the two predictions of BiSP are essentially identical frames, we regard them as complementary to each other, and thus $w_{f}$ and $w_{b}$ sum to one ($w_{f}+w_{b}=1$).

Based on the fusion error, we use the multi-scale anomaly evaluation method \cite{zhong2022bidirectional} for anomaly score calculation, which can be summarized as achieving a more comprehensive detection of anomalies at three different scales ($N=3$) through the error pyramid as shown in Eq.~(\ref{eq5}):

\begin{equation}\label{eq5}
\operatorname{PSNR}(I_{t}, \hat{I}_{t})=10 \log _{10}(\frac{1}{\sum_{i=0}^{N} v_{i}}), 
\end{equation}
where $v_{i}$ is the maximum prediction mean-square error based on the patch block in scale $i$ obtained by mean pooling. Finally, the PSNR is normalized to $[0,1]$ by Eq.~(\ref{eq6}), and smoothed by a Gaussian filter to produce the final anomaly score.

\begin{equation}\label{eq6}
S(I_{t})=\frac{\operatorname{PSNR}(I_{t}, \hat{I}_{t})-\min (\operatorname{PSNR}(I_{t}, \hat{I}_{t}))}{\max (\operatorname{PSNR}(I_{t}, \hat{I}_{t}))-\min(\operatorname{PSNR}(I_{t}, \hat{I}_{t}))}.
\end{equation}

\subsection{Variance Channel Attention}

The appearance disparity between most anomaly events and normal events is relatively significant, so most methods can easily cope with the above situations and achieve great anomaly scores, however, for a few ambiguous boundaries, most methods are unable to effectively deal with them again. For this reason, we propose a parallel structure for the variance channel attention (VarCA) mechanism, as shown in Fig.~\ref{fig2}(c). Because part of the anomalies are related to movement patterns, VarCA increases the feature variance between different events. During the testing phase, variance attention and channel attention focus on motion gaps and motion features of normal events, which increases the prediction errors for abnormal events. First, the variance of the spatial dimensions $D_{s}=H\times W$ on the different channels $C$ is computed after modeling the global feature $F_{a}\in R ^{B\times D_{s}\times 1}$ by 2D convolution with $1 \times 1$ filter sizes and a softmax activation function. Then, the input feature $F\in R ^{B\times C \times H\times W}$ are multiplied element-by-element with the results of variance attention map $M_{var}$ to get the variance feature $F_{var}\in R ^{B\times C \times H\times W}$. Meanwhile, the input features $F$ are also multiplied element-by-element with the results of channel attention map $M_{c}$ to get the channel feature $F_{c}\in R ^{B\times C \times H\times W}$, and finally, the two attention features are added element-by-element with the input features $F$ to obtain the variance channel feature $F_{vc}\in R ^{B\times C \times H\times W}$. The overview of VarCA is summarized as shown in Eq.~(\ref{eq7}):

\begin{equation}\label{eq7}
F_{vc}=F+F_{var}+F_{c},
\end{equation}

\begin{equation}\label{eq8}
\begin{aligned}
F_{var} & =F \otimes M_{var}(F) \\ & =\bar{F} \otimes \operatorname{softmax}\left(\left \| F_{a}-\frac{1}{D_{s}}\sum_{d=1}^{D_{s}}F_{a}^{d} \right \| _{2}^{2}\right),
\end{aligned}
\end{equation}

\begin{equation}\label{eq8.1}
F_{a}= \operatorname{softmax}(W^{1}(F)),
\end{equation}

\begin{equation}\label{eq9}
\begin{aligned}
F_{c} & = F \otimes M_{c}(F) \\ & = F \otimes (\delta W^{C}_{1}(\sigma W^{C/2}_{1}(P_{avg}(F)))),
\end{aligned}
\end{equation}
where $\delta$ and $\sigma$ denote the Sigmoid and ReLU activation functions, respectively. $\bar{F}\in R ^{B\times C \times D_{s}}$ denotes a tensor with the same data as $F$ but of a different shape. $P_{avg}$ denotes the global average-pooling operation, $W^{1}_{1}$, $W^{C}_{1}$ and $W^{C/2}_{1}$ denote the 2D convolution that are all $1 \times 1$ filter sizes and 1 channel, $C$ channel and $C/2$ channel, respectively.

\subsection{Context Spatial Attention}
Extracting spatio-temporal features is very important for anomaly detection, and prediction methods have better detection performance compared to reconstruction because they predict uninput video frames with certain temporal feature modeling capabilities. At the same time, for different event targets, their scale and location features have large differences, and the above features cannot be effectively extracted using basic convolution and pooling operations. Therefore, for the spatial features of different targets, we propose a serial structure for context spatial attention (ConSA) mechanism, as shown in Fig.~\ref{fig2}(d). First, we design a context feature extraction module $\mathrm{con(\cdot)}$, which uses four 2D dilation convolutions with different kernel sizes $k$ and expansion rates $r$ to extract the features and splice them into context feature $F_{t}\in R ^{B\times C_{t} \times H\times W}$, and $C_{t}=32 \times 4$ denotes the sum of the channels of the four dilation convolutions. Then, the context feature $F_{t}$ is multiplied element-by-element with the results of spatial attention map $M_{s}$ to get the spatial feature $F_{s}\in R ^{B\times C_{t} \times H\times W}$, and input to the 2D convolution with the same channels as the input feature $F\in R ^{B\times C \times H\times W}$ to get the final feature $F_{ts}\in R ^{B\times C \times H\times W}$. The overview of ConSA is summarized as shown in Eq.~(\ref{eq10}):

\begin{equation}\label{eq10}
\begin{aligned}
F_{ts} & = W^{C}_{1}(F_{s})  \\ &= W^{C}_{1} (F_{t} \otimes M_{s}(F_{t})),
\end{aligned}
\end{equation}

\begin{equation}\label{eq11}
F_{t} = \mathrm{con(\mathit{F})},
\end{equation}

\begin{equation}\label{eq12}
M_{s}= \delta (W^{1}_{3}(P_{avg}(F_{t}))),
\end{equation}
where $W^{C}_{1}$ denotes the 2D convolution with $1 \times 1$ filter sizes and the same channel as the $F$ channel, $W^{1}_{3}$ denotes the 2D convolution with $3 \times 3$ kernel sizes and 1 channel.

\section{Experiments} \label{sec4}

\subsection{Datasets and Implementation details}

To evaluate the qualitative and quantitative results of the proposed BiSP and to compare it with state-of-the-art algorithms, we conduct experiments on four benchmark unsupervised video anomaly detection datasets:

\noindent\textbf{UCSD Ped1 \& Ped2:} The Ped1 \& Ped2 \cite{wang2010anomaly} are the earliest proposed datasets. The Ped1 dataset contains a total of 70 video clips divided into 34 training clips and 36 testing clips, but due to the low pixel count of the video frames in Ped1, most VAD methods do not consider it as a benchmark dataset for the experiments. The Ped2 dataset contains 16 training clips and 12 testing clips, and its scenes are simple, and clear and have more distinct boundaries between normal and abnormal events, and the abnormal events are mostly biking, skateboarding, and driving, which are objects that are more distinct from humans.

\noindent\textbf{CUHK Avenue:} The Avenue \cite{lu2013abnormal} is a large-scale single-scene VAD dataset containing 16 training clips and 21 testing clips, with a total of 30,000 video frames and a large number of abnormal events related to human behavior, such as running, throwing a bag, and wrong direction.

\noindent\textbf{ShanghaiTech Campus:} The ShanghaiTech \cite{luo2017revisit} (Sh-Tech) dataset contains 13 scenes, more than 150 abnormal events such as running, loitering, jumping forward, etc., and up to 300,000 frames of surveillance video, making it the most challenging unsupervised VAD dataset. 

We conduct a large number of experiments using area under the curve (AUC) as the main evaluation metric. The proposed BiSP is trained using an Adam optimizer with a learning rate of 0.0002 and a cosine annealing scheduler to adjust the learning rate. The number of input frames $t$ is empirically set to 6 in the training phase and 7 in the testing phase. The resolution of all input images is rescaled to $256 \times 256$, and the pixel values are normalized to the range of $[-1,1]$. Following the assumptions of the unsupervised VAD \cite{pang2021deep}, all training sets are composed of video clips consisting of normal events collected from real scenes, whereas video clips containing anomaly events are defined as testing sets. The error weights $w_{f}$ and $w_{b}$ for Ped1, Ped2, Avenue, and Sh-Tech are set to (0.3, 0.7), (0.5, 0.5), (0.1, 0.9) and (0.7, 0.3), respectively.

\subsection{Comparison with State-of-the-Art Methods}

\begin{figure*}[ht]
	\centering
	\includegraphics[width=\textwidth]{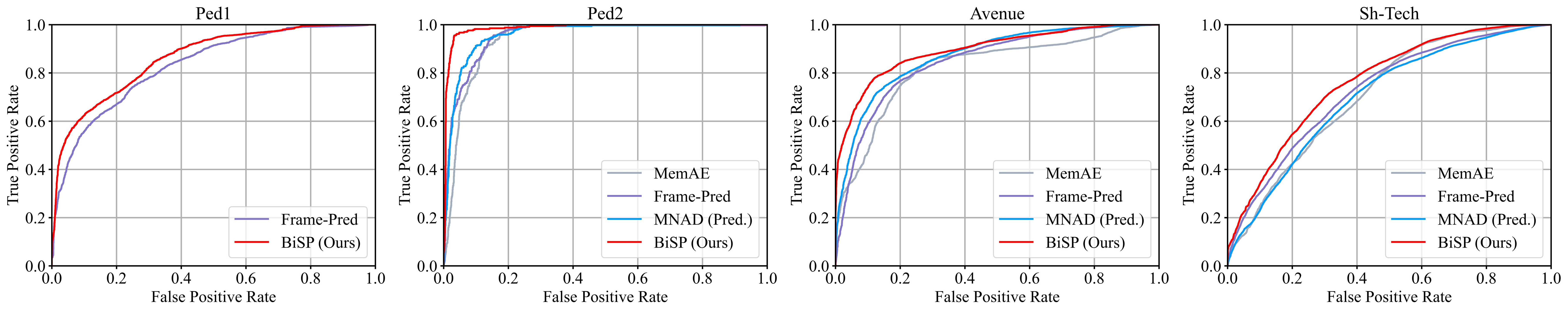}
	\caption{ROC curves of different methods on four benchmark datasets.} \label{fig5}
\end{figure*}

We compare the proposed BiSP with different AE-based reconstruction (Recon.) \cite{gong2019memorizing,park2020learning,chang2020clustering,fang2020multi,qiu2024video,lappas2024dynamic,kommanduri2024dast} and prediction (Pred.) \cite{liu2018future,park2020learning,hao2022spatiotemporal,huang2022self,yang2022dynamic,zhang2022hybrid,cheng2023spatial,liu2023msn,huang2022boosting,yang2024context,qiu2024video,zhang2022mutual,fang2020anomaly,zhong2022bidirectional,li2023multi} methods on four benchmark datasets. The AUC performances of the different methods are shown in Table~\ref{tab1}. It can be seen that BiSP outperforms the other methods on both Ped2 and Sh-Tech datasets. Fig.~\ref{fig5} shows the frame-level ROC curves comparing BiSP with the baseline method, where a larger area under the curve indicates a higher detection accuracy.

Reconstruction methods have outputs that are consistent with the inputs, thus making it difficult to learn temporal features, causing the performance of such methods to fall behind most prediction methods. However, we find that a few reconstruction methods \cite{wang2023memory,lappas2024dynamic,kommanduri2024dast} perform well in Avenue. This is because there are phenomena such as uneven background illumination in Avenue, which leads to the prediction methods not being able to focus on generating high-quality prediction frames, while the reconstruction methods can cope with this problem well. This paper focuses on comparing prediction methods, which are further categorized into single-frame prediction (Single-Pred.) and dual-frame prediction (Dual-Pred.).

\begin{table}
	\centering
	\caption{Comparative results of frame-level AUC (\%) for the benchmark datasets. Best and second best performance are highlighted in bold and underlined, respectively.}\label{tab1}
	\begin{tabular}{l|l|l|l|l|l}
		\hline
		& Method & Ped1 & Ped2 & Avenue & Sh-Tech \\
		\hline
		\multirow{8}{*}{\rotatebox{90}{Recon.}} & MemAE \cite{gong2019memorizing} & – & 94.1 & 83.3 & 71.2 \\
		& MNAD \cite{park2020learning} & – & 90.2 & 82.8 & 69.8 \\
		& Cluster-AE \cite{chang2020clustering} & – & 96.5 & 86.0 & 73.3 \\
		& MESDnet \cite{fang2020multi} & – & 95.6 & 86.3 & 73.2 \\
		& MAAM-Net \cite{wang2023memory} & – & 97.7 & \textbf{90.9} & 71.3 \\
		& Qiu \emph{et al.} \cite{qiu2024video} & – & 90.8 & 83.1 & 73.3 \\
		& C3DSU with DDL \cite{lappas2024dynamic} & – & \underline{98.5} & \underline{90.4} & 74.3 \\
		& DAST-Net \cite{kommanduri2024dast}  & 85.4 & 97.9 & 89.8 & 73.7 \\
		\hline
		\multirow{12}{*}{\rotatebox{90}{Single-Pred.}}& Frame-Pred \cite{liu2018future} & 83.1 & 95.4 & 85.1 & 72.8 \\
		& MNAD \cite{park2020learning} & 81.1 & 97.0 & 88.5 & 70.5 \\
		& STCEN \cite{hao2022spatiotemporal}  & 82.5 & 96.9 & 86.6 & 73.8 \\
		& SSAGAN \cite{huang2022self}  & 84.2 & 96.9 & 88.8 & 74.3 \\
		& DLAN-AC \cite{yang2022dynamic}  & – & 97.6 & 89.9 & 74.7 \\
		& Zhang \emph{et al.} \cite{zhang2022hybrid} & 85.2 & 95.8 & 84.9 & 71.4 \\
		& STGCN-FFP \cite{cheng2023spatial} & – & 96.9 & 88.4 & 73.7 \\
		& ASTNet \cite{le2023attention} & – & 97.4 & 86.7 & 73.6 \\
		& MSN-net \cite{liu2023msn}  & – & 97.6 & 89.4 & 73.4 \\
		& VADNet \cite{huang2022boosting}  & – & 96.8 & 87.3 & \underline{75.2} \\
		& Qiu \emph{et al.} \cite{qiu2024video} & – & 92.2 & 86.2 & 73.8 \\
		& Trinity \cite{yang2024context} & – & 97.9 & 88.5 & 74.1 \\
		
		\hline
		\multirow{6}{*}{\rotatebox{90}{Dual-Pred.}}& MLIPN \cite{zhang2022mutual} & 83.9 & 96.0 & 85.9 & 73.1 \\
		& SIGnet \cite{fang2020anomaly} & \underline{86.0} & 96.2 & 86.8 & – \\
		& DEDDnet (Fusion) \cite{zhong2022bidirectional} & – & 98.1 & 89.0 & 74.5 \\
		& DEDDnet (Parallel) \cite{zhong2022bidirectional} & – & 97.4 & 88.6 & – \\		
		& MGAN-CL \cite{li2023multi}  & – & 96.5 & 87.1 & 73.6 \\
		& BiSP (Ours) & \textbf{86.3} & \textbf{98.6} & 89.5 & \textbf{76.4} \\
		\hline
	\end{tabular}
	
\end{table}

Compared with the prediction methods, in Ped2, the boundary between normal and anomalies is sufficiently obvious, and the proposed BiSP can further expand the boundary of different events on this basis, thus improving the detection performance. Among the dual-frame prediction methods, MLIPN \cite{zhang2022mutual} achieves two results for unidirectional frame prediction through two independent generators, and the remaining four methods are bidirectional frame prediction, whose testing phases are to co-predict the same intermediate frames through the video frames at both ends, which is considered to be a simple and efficient detection way. SIGnet \cite{fang2020anomaly} and MGAN-CL \cite{li2023multi} both use the AE as the generator, which is a much easier and more efficient detection mode than using only the AE as the primary network. GAN with better feature representation than DEDDnet \cite{zhong2022bidirectional} and BiSP which only use AE as the primary network. However, this is because DEDDnet and BiSP enhance the model's attention to normal events by designing different attention mechanisms, thus guaranteeing that some of the ambiguous abnormal events are detected. Here we compare the Fusion and Parallel versions of DEDDnet \cite{zhong2022bidirectional}, the Parallel version is similar to the proposed BiSP, i.e., it does not fuse the bidirectional prediction information and implements two independent frame prediction methods. Since the detection results at Sh-Tech are not provided for the parallel version, this paper compared to the fusion version, the BiSPincreases by 1.9 \%. Although these two results are approximate, the ability of our method to learn features independently of forward and backward predictions indicates that our proposed method is more robust.

\subsection{Ablation Study}

\subsubsection{Error weights hyperparameter analysis}

\begin{figure}
	\centering
	\includegraphics[width=3.5in]{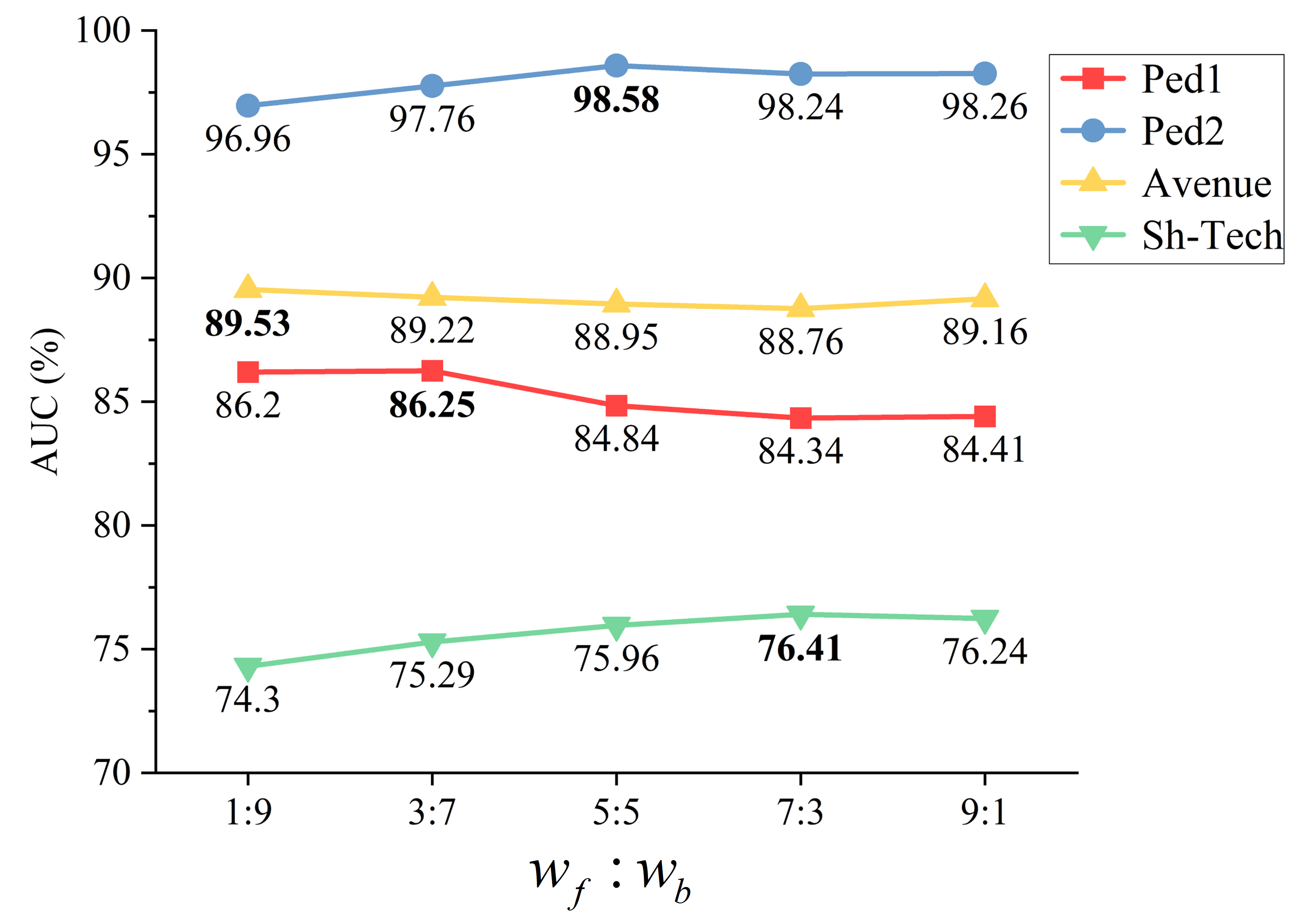}
	\caption{Comparison of AUC for different error weight hyperparameters, bolded for optimal performance.} \label{fig6.0}
\end{figure}

We discuss the effect of different error weights hyperparameters for the AUC and perform several experiments on four datasets. As shown in Fig.~\ref{fig6.0}, we use five sets of hyperparameters to measure the performance of different error weights, where the horizontal axis represents the relationship between $w_{f}$ and $w_{b}$ in terms of proportions, i.e., $3:7$ means that $w_{f}$ is  0.3 and $w_{b}$ is 0.7. It can be intuitively seen that the AUC changes obtained from different error weights do not fluctuate much, and in particular, there is only a 0.78\% fluctuation in Avenue. Although such fluctuations are acceptable in other datasets, how to choose appropriate error weight ratios for different datasets to optimize the AUC, is also a problem that needs to be considered by all hybrid methods.

\subsubsection{Component and training strategy analysis}
We conduct ablation studies to measure the effectiveness of skip-frame (SkipF) and the performance gained by each attention. The ablation study takes the Ped2 and Avenue datasets as an example and verifies the effectiveness by combining the proposed attentions one by one, and lastly the Model 1 and Model 5 validate SkipF. The results of the ablation studies are shown in Table~\ref{tab2}. To safeguard the consistency of the ablation studies, the error weight hyperparameters $w_{f}$ and $w_{b}$ in this subsection were both set to 0.5.

\begin{table}[ht]
	\centering
	\caption{Ablation studies of the proposed BiSP on Ped2 and Avenue datasets.}\label{tab2}
	\begin{tabular}{l|l|l|l|l|l}
		\hline
		Model & SkipF & VarCA & ConSA & Ped2 & Avenue  \\
		\hline
		1 & \usym{2717} & \usym{2717} & \usym{2717} & 97.0 & 85.0  \\
		2 & \usym{2714} & \usym{2717} & \usym{2717} & 97.5 & 85.8 \\
		3 & \usym{2714} & \usym{2714} & \usym{2717} & 98.2 & 87.0  \\
		4 & \usym{2714} & \usym{2717} & \usym{2714} & 98.0 & 87.1  \\
		5 & \usym{2717} & \usym{2714} & \usym{2714} & 98.0 & 87.3  \\
		6 & \usym{2714} & \usym{2714} & \usym{2714} & 98.6 & 89.0   \\
		\hline
	\end{tabular}
\end{table}

It can be seen from Table~\ref{tab2}, that the variant Models 2, 3, and 4 reduce the detection accuracy. The attention mechanism has less effect on ped2 due to its clear object and large boundary disparity, meanwhile, skip frames enhance the motion features of the objects. In contrast, in the more complex Avenue dataset, the effect of attention is more significant. As Model 6 shows, when adding two attentions in Model 2, the AUC performance significantly improved by 1.1\% and 3.2\%, respectively, and combining the results of Models 3 and 4 revealed that employing only single attention may cause abnormal generalization and reduce performance. Compared to Models 2 and 6, Models 1 and 5 use consecutive frames as model inputs instead of SkipF, resulting in a slight decrease in performance. This is because skip frames enhance the extraction efficiency of motion features. Anomaly events in the existing dataset are heavily associated with fast motion, and thus the model trained using the skip frame works best. Meanwhile, as shown in Model 1, using bidirectional prediction without any components still achieves better performance. This is due to the ability of the forward and backward prediction frames to catch the anomaly error simultaneously, which maximizes the detection efficiency.

Furthermore, to explore the effectiveness of the proposed method under different training strategies, this paper also designs three additional ablation models, i.e., "Forward", "Backward" and "Fusion". Table~\ref{tab3} shows the AUC performance of different training strategies in different datasets. "Forward" or "Backward" denotes the use of forward skip frames or backward skip frames for training and testing only in the framework of single AE, respectively. "Fusion" denotes that based on the proposed BiSP framework, the input features of the decoder are transformed from only the separate output features of the forward/backward encoder to the fusion of the low-dimensional hidden features of the dual-AE. Here, we use simple element-wise addition to represent the feature fusion process. To target the evaluation of the model and the calculation of the AUC, the single AE model only needs to calculate the error from the corresponding forward or backward frames. In contrast, the testing results of the dual-AE model are represented as the same frames predicted from the forward and backward directions, so the average of the two predicted frames is used to calculate the error.

\begin{table}[h]
	\centering
	\caption{Frame-level AUC comparison of anomaly scores under different feature learning methods.}\label{tab3}
	\begin{tabular}{l|l|l|l|l}
		\hline
		& Ped1 & Ped2 & Avenue & SH-Tech   \\
		\hline
		Forward & 82.9 & 96.3 & 86.4 & 73.6  \\
		Backward & 83.4 & 96.6 & 86.8 & 73.4 \\
		Fusion & 83.9 & 98.0 & 88.5 & 74.6 \\
		BiSP & 84.8 & 98.6 & 89.0 & 76.0  \\
		
		\hline
	\end{tabular}
\end{table}

From Table~\ref{tab3}, we can note that the AUC of the "Forward" and "Backward" variant models for a single AE is lower than that of the model with dual-AE, which indicates that the bidirectional training strategy outperforms the unidirectional. Furthermore, the "Fusion" model is yet lower than the proposed method, which is different from the conclusion of DEDDnet \cite{zhong2022bidirectional}. This is because DEDDnet uses an identical video frame for bidirectional training, while the prediction frame does not overlap with the training video frame. In contrast, the bidirectional training video frames of BiSP do not overlap and the forward training video frames contain the prediction frame information of the backward AE, as does the backward training. This cross-fusion disrupts the feature extraction process of the prediction network and leads to a slight decrease in performance.

\begin{figure}[h]
	\centering
	\includegraphics[width=\linewidth]{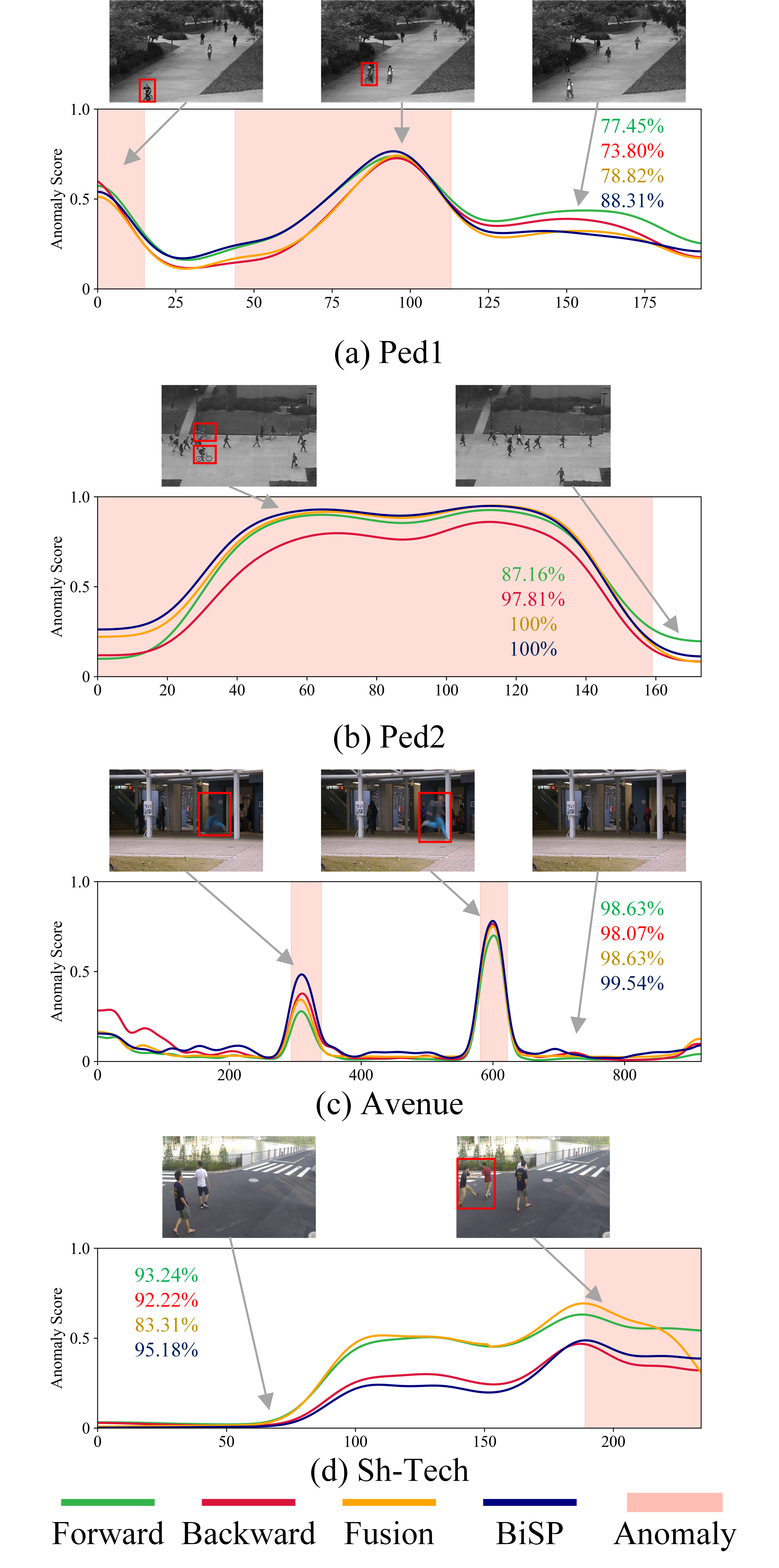}
	\caption{Anomaly score curves and scores for several testing video clips of different methods on four benchmark datasets. The different color curves and numbers represent the score curves and scores of different models. The area marked by the red border contains abnormal events.} \label{fig6.1}
\end{figure}

To specifically analyze the discrimination ability of the different models for anomaly and normal, Fig.~\ref{fig6.1} shows its anomaly curves and scores in the testing video clips. Anomaly scores are obtained based on different thresholds, which depend on the false and true positive rates. We categorized each testing video frame by normalized anomaly scores and thresholds, i.e., a score close to 0 indicates normal, and close to 1 indicates an anomaly. The anomaly in the testing video frames in Fig.~\ref{fig6.1} is more obvious, so the curves trend similarly for the different methods, but there is still a slight difference. For instance, the dual-AE model in Fig.~\ref{fig6.1}(a) has lower scores in the normal interval than the single AE model, while performing higher in the anomaly interval, which further demonstrates the advantage of bidirectional prediction. In Fig.~\ref{fig6.1}(d), the plunge of the fusion model's score in the anomaly interval is caused by the cross information interfering with the prediction model, which results in a much lower performance than the proposed method.

\subsection{Visualization Results}

\begin{figure*}[ht]
	\centering
	\includegraphics[width=\textwidth]{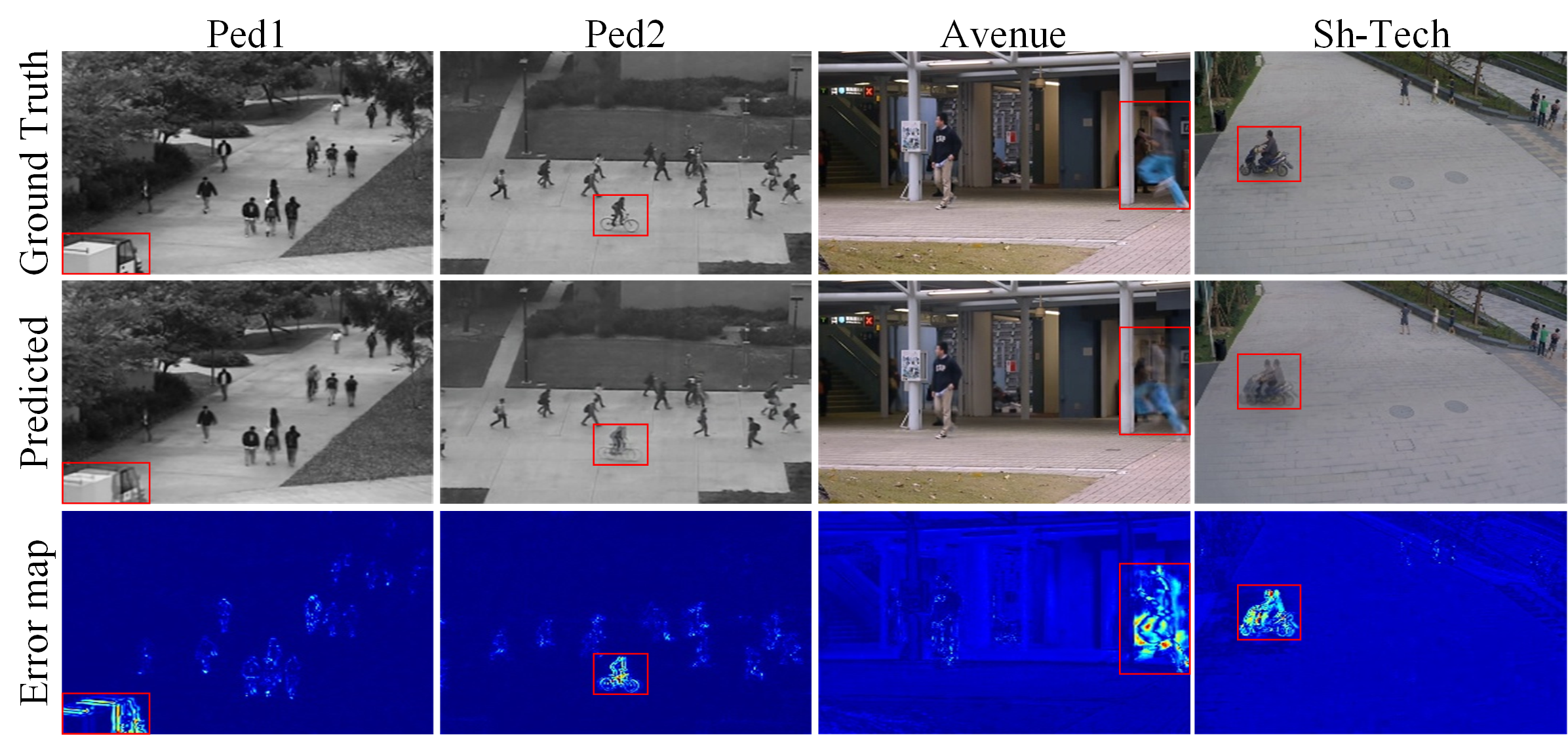}
	\caption{Visualization results for four datasets: Ground Truth (top), Predicted frames (middle), and Error maps (bottom), where brighter colors represent larger errors. The area marked by the red border contains abnormal events.} \label{fig6}
\end{figure*}
Fig.~\ref{fig6} shows the visualization results of the proposed BiSP in different datasets and marks the exact location of the anomalies. It can be noticed that due to the bidirectional prediction, the abnormal events demonstrate overlapping blurred states, especially for abnormal events with significant boundaries, the brightness of the prediction error maps is very obvious, meanwhile, the errors of the normal events are negligible. In the Avenue dataset, in addition to the obvious predictions that can be observed, some of the backgrounds are found to have slight errors than the surrounding information. Although Avenue belongs to a single scene, it can find the phenomenon of background bias by observing the training set, which shows different degrees of light intensities in various video clips.

\begin{figure*}[b]
	\centering
	\includegraphics[width=\textwidth]{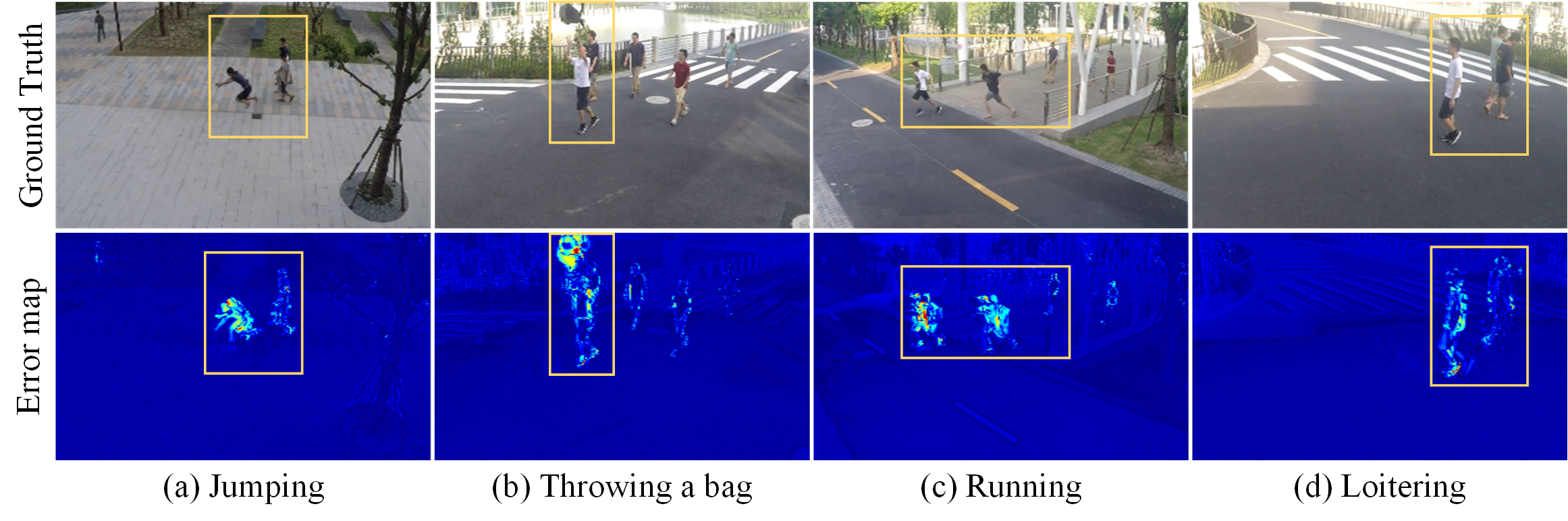}
	\caption{Examples of ambiguous boundary in the Sh-Tech dataset: Ground Truth (top) and Error maps (bottom). The area marked by the yellow border contains normal and abnormal events.} \label{fig7}
\end{figure*}

In contrast to the clear boundary shown in Fig.~\ref{fig6}, Fig.~\ref{fig7} illustrates the ambiguous boundary in the Sh-Tech dataset, including four anomaly events, i.e., jumping, throwing a bag, running, and loitering. It can be seen that ambiguous abnormal events are distinguished only by human behavior. In the error maps, the proposed BiSP achieves efficient discrimination of the ambiguous anomalies in Fig.~\ref{fig7}(a)-(c), with a significant gap between normal and anomaly. For the loitering behavior in Fig.~\ref{fig7}(d), the white-dressed human is much less morphological differences from the others, such anomalies in this behavior are more difficult to detect. Still, BiSP achieves the desired discriminative ability by widening the intra-domain disparity between different events.

\section{Conclusion} \label{sec5}

In this paper, we propose a bidirectional skip-frame prediction network for unsupervised VAD, called BiSP. Based on a dual-stream autoencoder, the variance channel attention with the parallel structure and the context spatial attention with the serial structure are constructed to enlarge the intra-domain disparity between normal and anomaly events and improve the model's detection performance, especially in ambiguous boundaries. Experimental results on benchmark datasets and the ablation study validate the performance of our method as superior or equivalent to state-of-the-art methods. In future work, we aim to focus on semantic information for different scenes, which guarantees, at least in principle, that the diverse anomalies are better defined.

\section{Acknowledgments}

This research was funded by Natural Science Foundation of Shaanxi Province, China(2024JC-ZDXM-35, 2024JC-YBMS-458, 2024JC-YBMS-573), National Natural Science Foundation of China (No.52275511) and Young Talent Fund of Association for Science and Technology in Shaanxi, China(20240146).

\nocite{*}
\bibliographystyle{IEEEtran}
\bibliography{references.bib}

\end{document}